\documentclass[extended-abstract]{ccn}

\title{Focus particles and scalar inferences across humans and language models}

\author{Catherine M. Brousse and Nelu D. Radpour \\
  Department of Psychology, Florida State University \\
  \email{\{brousse,radpour\}@psy.fsu.edu}}
\begin{document}

\maketitle

\begin{abstract}
We compared scalar judgments from humans and a large language model for sentences containing the focus particles \textit{even} and \textit{only} across four response-scale configurations varying in spatial format and label mapping. Both humans and the LLM showed stable differences between \textit{even} and \textit{only} across configurations, suggesting that scalar judgments are driven primarily by semantic structure rather than spatial organization. Although the model reproduced aggregate behavioral patterns, its lack of response variability raises questions about the suitability of LLMs as models of human cognition.
\end{abstract}

\section{Introduction}
Understanding how humans and artificial systems interpret structured meaning is a central question in cognitive science and natural language processing. Focus-sensitive particles such as \textit{even} and \textit{only} provide a clear case of how small function elements can systematically reshape interpretation of the sets of alternatives under consideration. These expressions do not change the core propositional content of a sentence but instead modify its inferred meaning by introducing expectations about what is surprising, typical, or exclusive. For example, \textit{even} tends to highlight unexpected or low-probability alternatives, while \textit{only} introduces exclusivity by restricting the set of valid alternatives.

However, it remains unclear whether the resulting scalar evaluations reflect spatially organized magnitude representations or more general, valence-based mechanisms. One account proposes that scalar judgments are grounded in spatial coding, such that abstract magnitudes are mapped onto spatial dimensions like left-right or up-down space \parencite{dehaene1993mental}. On this view, the spatial configuration of response options should systematically influence judgments. An alternative account derives from the polarity correspondence principle, which proposes that performance is facilitated when the conceptual polarity of a stimulus aligns with the polarity of response mappings \parencite{proctor2006polarity}. From this perspective, differences between \textit{even} and \textit{only} should remain largely stable across spatial configurations because judgments are driven primarily by evaluative and linguistic structure rather than spatial magnitude representations.

Recent work suggests that LLMs develop structured internal representations of linguistic information \parencite{kryvosheieva2025different}, raising the question of whether these systems exhibit behavior patterns similar to those of humans. In this study, we examine whether humans and a language model assign similar ability judgments when interpreting sentences containing \textit{even} and \textit{only} under different response scale configurations. Across conditions, we manipulate both scale orientation and spatial format to test whether scalar judgments are systematically influenced by spatial alignment or by linguistic and evaluative structure. Unlike humans, whose representations of space are shaped by embodied interactions with the physical world, transformer-based language models process textual input tokens simultaneously through bidirectional attention \parencite{vaswani2017attention}, so the spatial orientation of a response scale may not map onto directional representations the way it does for humans. Comparing human and model responses therefore provides a test of whether similar scalar judgments emerge from shared representational principles or from different underlying mechanisms.

\section{Method} 
We created 60 sentences that describe people performing ability-relevant tasks (e.g., "Mike can bake a cake.”). Thirty items contained the particle \textit{even} and another thirty items contained \textit{only}. Human participants ($N = 108$) completed one of four Qualtrics studies. We manipulated the spatial configuration of the response scale along two dimensions: spatial format (horizontal vs. vertical) and label mapping (standard vs. reversed). In the standard mapping condition, low ability appeared on the left (or bottom in vertical layouts), and high ability appeared on the right (or top in vertical layouts). In the reversed mapping conditions, these labels were flipped. These manipulations yielded four experimental conditions: horizontal-standard, horizontal-reversed, vertical-standard, and vertical-reversed. On each trial, participants read a target sentence and rated the individual on a 5-point Likert scale.

The language model data were collected using Llama 3.3 70B \parencite{touvron2023llama}, a 70-billion parameter open-source autoregressive language model, accessed via the Groq API (model string: \texttt{llama-3.3-70b-versatile}). Each stimulus sentence was presented in a prompt mirroring the structure of the human survey, including the response scale configurations. The model was instructed to respond with a single integer from 1 to 5 based on the implied ability of the person described. To sample from the model's response distribution, each prompt was run 20 times at a temperature of 1.0, yielding 4,800 observations across 60 sentences and 4 scale conditions.

\section{Results}
Ratings were analyzed using a linear mixed-effects model with Particle (Even vs. Only), Source (Human vs. LLM), Format (Horizontal vs. Vertical), and Mapping (Standard vs. Reversed) as fixed effects and random intercepts for participants and items. The model revealed a significant effect of Particle ($\beta = 2.20$, $SE = 0.09$, $p < .001$) and a significant Particle × Source interaction ($\beta = 1.38$, $SE = 0.03$, $p < .001$). Human ratings averaged 2.07 for \textit{even} sentences and 4.31 for \textit{only} sentences, whereas the LLM produced more extreme ratings (1.41 and 5.00, respectively). Follow-up comparisons of estimated marginal means showed that both humans and the LLM assigned higher ratings to \textit{only} sentences than to \textit{even} sentences ($p < .001$). However, the difference between particles was larger for the LLM (3.58) than for humans (2.20).

These findings suggest that both humans and the LLM were sensitive to the scalar implications associated with \textit{even} and \textit{only}. Across conditions, \textit{even} sentences were associated with lower ability judgments, whereas \textit{only} sentences were associated with higher ability judgments. Critically, changing the orientation or label mapping of the response scale did not eliminate or reverse this pattern, suggesting that scalar judgments were driven primarily by the semantic and evaluative content of the particles rather than by the spatial arrangement of the response scale. This pattern provides stronger support for an evaluative account of scalar judgments than for an account based on spatial coding. At the same time, the LLM produced more extreme judgments than human participants. In particular, \textit{only} sentences were assigned the maximum rating across conditions for every observation, indicating that the model treated the implications of \textit{only} more categorically than humans, who exhibited more graded responses.
\begin{figure}[t]
    \centering
    \includegraphics[width=1\linewidth]{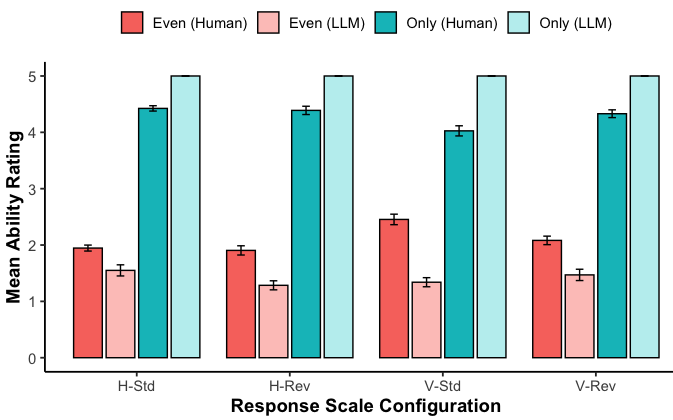}
    \caption{Human and LLM ability ratings by particle and response-scale configuration. Error bars: 95\% CIs; absent for the LLM in \textit{only} conditions as all responses were identical.}
    \label{fig:placeholder}
\end{figure}
\section{Discussion}
Although the LLM reproduced the qualitative human pattern, it diverged from human behavior in two ways. First, its judgments were more extreme, assigning the scale maximum rating on 100\% of \textit{only} trials while exhibiting substantially less variability than human participants for \textit{even} trials as well. Second, despite a sampling temperature of 1.0 (often assumed to produce human-like response variability), repeated sampling produced highly consistent responses. Temperature can only add randomness to the distribution of responses the model already has; when nearly all of the model's probability is concentrated on a single response, repeated sampling will tend to return the same response.

Consistent with this account and following prior work \parencite{rodriguez2026modeling, qiu2025can}, a follow-up temperature sweep on a subset of
items showed that model responses remained highly stable across
sampling temperatures from 0.0 to 2.0 (20 samples per item, condition, and
temperature), in both standard and reversed scale mappings. This pattern indicates that increasing temperature alone was insufficient to produce human-like response variability. Sampling a model many times is therefore not equivalent to sampling many human participants; variability across runs of one model reflects noise around a single fixed judgment, whereas variability across people reflects genuine individual differences in interpretation. Indeed, work has shown that selecting an appropriate sampling
temperature remains an open challenge \parencite{du2025optimizing,
li2026necessity}.

The absence of variability in our simulations further raises questions about the suitability of current LLMs as models of human cognition, given that individual differences are a defining feature of human behavior. More broadly, this is consistent with recent findings that contemporary LLMs fail to reproduce the breadth of variability observed in human populations, even in cases where average task performance and aggregate response patterns are well matched \parencite{qiu2025can}.

\printbibliography

\end{document}